\documentclass[journal]{IEEEtran}
\usepackage{algorithm} 
\usepackage[numbers]{natbib}
\usepackage{multicol}
\usepackage[colorlinks=true]{hyperref}

\usepackage{graphicx}
\usepackage{color}
\usepackage{booktabs}

\usepackage{enumitem}
\usepackage[noend]{algpseudocode}
\algrenewcommand\alglinenumber[1]{\normalsize #1:}
\usepackage{amsmath}

\newcommand{\ie}{\textit{i}.\textit{e}., }
\newcommand{\eg}{\textit{e}.\textit{g}. }
\newcommand{\rpm}{\raisebox{.2ex}{$\scriptstyle\pm$}}
\newcommand{\mypara}[1]{\par\vspace*{0.8mm}\noindent\textbf{\textit{#1}}}

\begin{document}
%
\title{TossingBot: Learning to Throw Arbitrary Objects\\with Residual Physics}
%
%
%

\author{
Andy Zeng$^{1,2}$, 
Shuran Song$^{1,2,3}$, 
Johnny Lee$^{2}$, 
Alberto Rodriguez$^{4}$, 
Thomas Funkhouser$^{1,2}$ 
\vspace{1mm} \\ 
$^{1}$Princeton University\quad
$^{2}$Google\quad
$^{3}$Columbia University\quad
$^{4}$Massachusetts Institute of Technology
\vspace{1mm} \\
\href{http://tossingbot.cs.princeton.edu/}{http://tossingbot.cs.princeton.edu}
\vspace{-2em}

\thanks{
This journal paper is a revision of a conference paper appearing in Robotics: Science and Systems (RSS) 2019 \cite{zeng2019tossingbot}.
}
}

%
%

\markboth{IEEE TRANSACTIONS ON ROBOTICS}%
{Shell \MakeLowercase{\textit{et al.}}: Bare Demo of IEEEtran.cls for IEEE Journals}
%



\maketitle

\begin{abstract}

We investigate whether a robot arm can learn to pick and throw arbitrary rigid objects into selected boxes quickly and accurately. Throwing has the potential to increase the physical reachability and picking speed of a robot arm.
However, precisely throwing \textit{arbitrary objects} in unstructured settings presents many challenges: from acquiring objects in grasps suitable for reliable throwing, to handling varying object-centric properties (\eg mass distribution, friction, shape) and complex aerodynamics. In this work, we propose an end-to-end formulation that jointly learns to infer control parameters for grasping and throwing motion primitives from visual observations (RGB-D images of arbitrary objects in a bin) through trial and error. Within this formulation, we investigate the synergies between grasping and throwing (\ie learning grasps that enable more accurate throws) and between simulation and deep learning (\ie using deep networks to predict residuals on top of control parameters predicted by a physics simulator). The resulting system, \textbf{\textit{TossingBot}}, is able to grasp and successfully throw arbitrary objects into boxes located outside its maximum reach range at 500+ mean picks per hour (600+ grasps per hour with 85\% throwing accuracy); and generalizes to new objects and target locations. Videos are available at \href{http://tossingbot.cs.princeton.edu/}{http://tossingbot.cs.princeton.edu}


\end{abstract}

\begin{IEEEkeywords}
robotic manipulation, deep learning, perception
\end{IEEEkeywords}

%
\IEEEpeerreviewmaketitle

\section{Introduction}
\label{sec:introduction}

\begin{figure}[t]
\centering
  \vspace{2mm}
  \includegraphics[width=\linewidth]{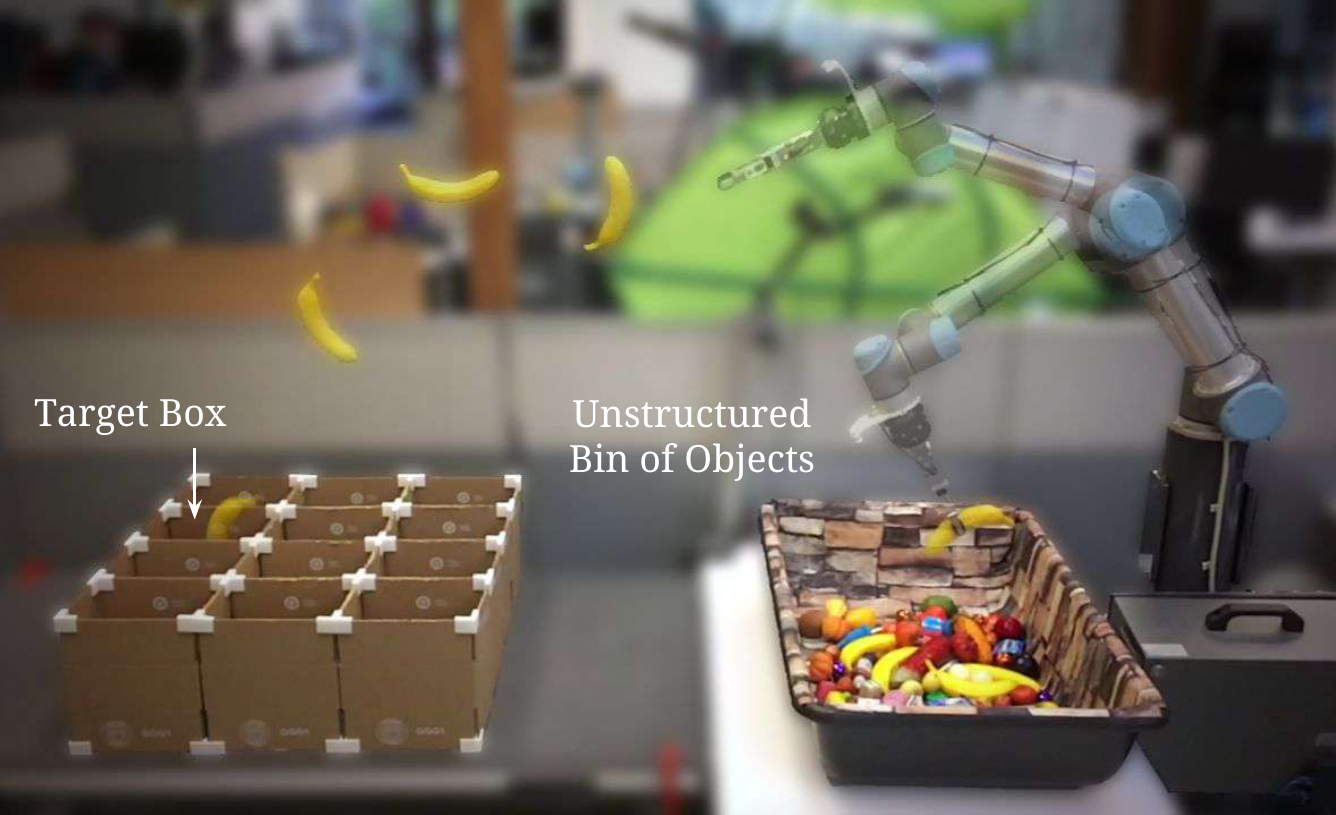}
  \caption{\textbf{TossingBot} learns to grasp arbitrary objects from an unstructured bin and to throw them into target boxes located outside its maximum kinematic reach range. The aerial trajectory of different objects are controlled by jointly optimizing grasping policies and throwing release velocities.} 
  \label{fig:teaser}
  \vspace{-1em}
\end{figure}

\IEEEPARstart{T}{hrowing} is a means to increase the capabilities of a manipulator by exploiting dynamics, a form of dynamic extrinsic dexterity~\cite{dafle2014extrinsic}. In the case of pick-and-place, throwing enables a robot arm to place objects rapidly into boxes located outside its maximum kinematic range, which not only reduces the total physical space used by the robot, but also maximizes its picking efficiency. Rather than having to transport objects to their destination before executing the next pick, objects are instead immediately ``passed to Newton'' (see Fig. \ref{fig:teaser}).

\begin{figure}[t]
\centering
  \includegraphics[width=\linewidth]{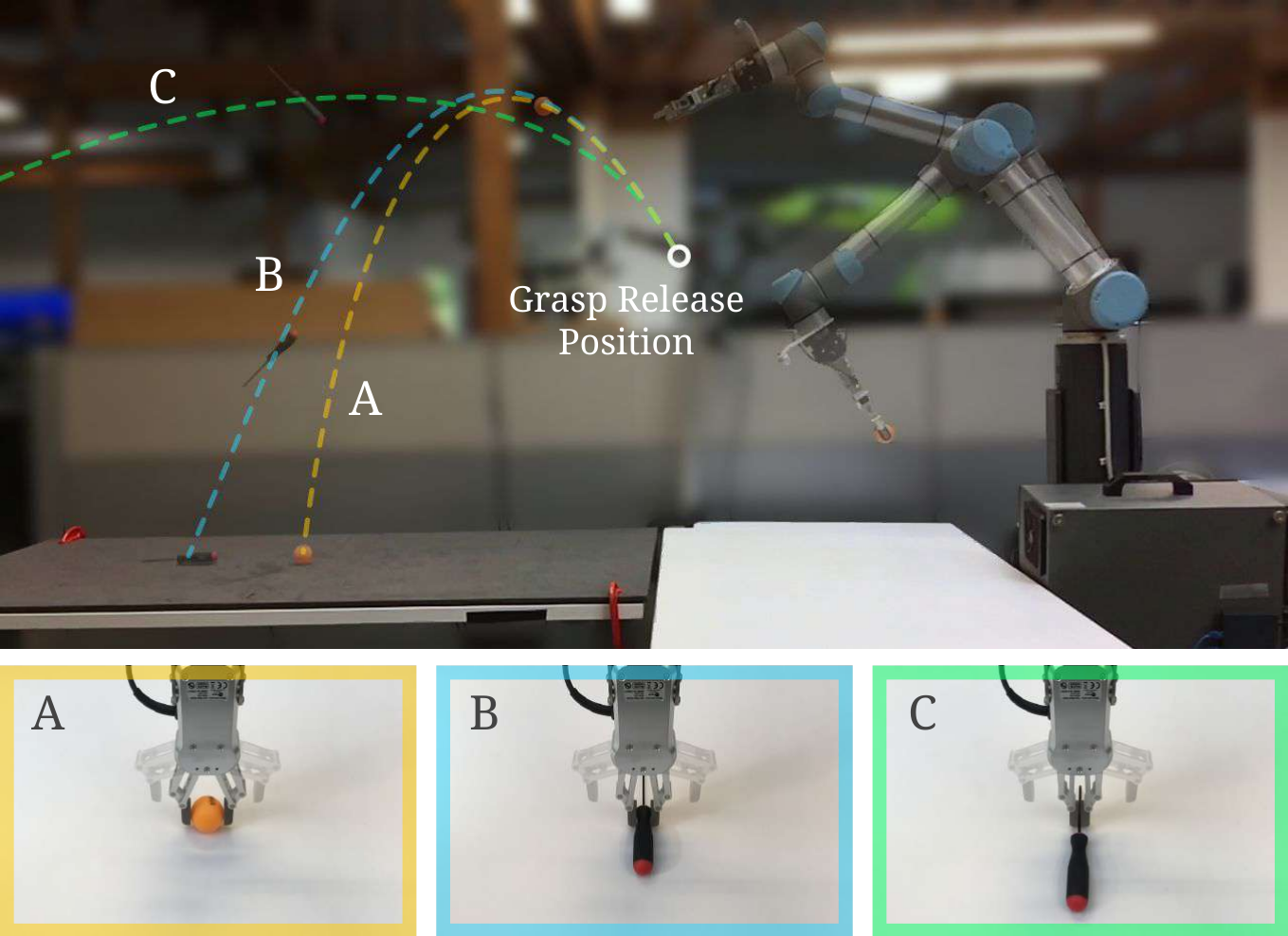}
  \caption{\textbf{Projectile trajectories} of a thrown ping pong ball (a), screwdriver grasped and thrown by its handle (b), and the same screwdriver grasped and thrown by its shaft (c). The difference between (a) and (b) is largely due to aerodynamics, while the difference between (b) and (c) is largely due to grasping at different offsets from the object's center of mass (near the handle). Our goal is to learn joint grasping and throwing policies that can compensate for these differences to achieve accurate targeted throws.}
  \label{fig:grasping-affects-tossing}
  \vspace{-1em}
\end{figure}

However, precisely throwing \textit{arbitrary objects} in unstructured settings is challenging because it depends on many factors: from pre-throw conditions (\eg initial grasp of the object) to varying object-centric properties (\eg mass distribution, friction, shape) and dynamics (\eg aerodynamics). For example, grasping a screwdriver near the tip before throwing it can cause centripetal accelerations to swing it forward with significantly higher release velocities -- resulting in drastically different projectile trajectories than if it were grasped closer to its center of mass (see Fig. \ref{fig:grasping-affects-tossing}). Yet regardless of how it is grasped, its aerial trajectory would differ from that of a thrown ping pong ball, which can significantly decelerate after release due to air resistance. Many of these factors are notoriously difficult to model or measure analytically \cite{mason1993dynamic,lynch1999dynamic} -- hence prior studies are often confined to assuming homogeneous pre-throw conditions (\eg object fixtured in gripper or manually reset after each throw) with predetermined, homogeneous objects (\eg balls or darts). Such assumptions rarely hold in real unstructured settings, where a throwing system needs to acquire its own pre-throw conditions (via grasping) and adapt its throws to account for varying properties and dynamics of arbitrary objects. 

In this work, we present \textbf{\textit{TossingBot}}, an end-to-end formulation that uses trial and error to learn how to plan control parameters for grasping and throwing from visual observations. The formulation learns grasping and throwing jointly -- discovering grasps that enable accurate throws, while learning throws that compensate for the dynamics of arbitrary objects. There are two key aspects to our system:
\begin{itemize}[leftmargin=*]
    \item \textbf{Joint learning of grasping and throwing policies} with a deep neural network that maps from visual observations (of objects in a bin) to control grasping and throwing parameters: the likelihood of grasping success for a dense pixel-wise sampling of end effector orientations and locations \cite{zeng2018learning}, and the throwing release velocities for each sampled grasp. Grasping is directly supervised by the accuracy of throws (grasp success = accurate throw), while throws are directly conditioned on specific grasps (via dense predictions). As a result, the end-to-end policy learns to execute stable grasps that lead to predictable throws, as well as throwing velocities that account for the variations in object-centric properties and dynamics that can be inferred from visual information.
    \item \textbf{Residual learning of throw release velocities} $\delta$ on top of velocities $\hat{v}$ predicted by a physics controller based on an ideal ballistic motion. The complete controller uses the superposition of the two predictions to obtain a final throwing release velocity $v=\hat{v}+\delta$. The physics-based controller uses ballistics to provide consistent estimates of $\hat{v}$ that generalize well to different landing locations, while the data-driven residuals learn to exploit those grasps, and compensate for object-centric properties and dynamics. Our experiments show that this hybrid data-driven method, \textit{Residual Physics}, leads to significantly more accurate throws than baseline alternatives.
\end{itemize}
This formulation enables our system to grasp and throw arbitrary objects reliably into target boxes located outside its maximum reach range at 500+ mean picks per hour (MPPH), and generalizes to new objects and target landing locations.

The primary contribution of this paper is to provide new perspectives on throwing: in particular -- its relationship to grasping, its efficient learning by combining physics with trial and error, and its potential to improve practical real-world picking systems. We provide several experiments and ablation studies in both simulated and real settings to evaluate the key components of our system. We observe that throwing performance strongly correlates with the quality of grasps, and experimental results show that our formulation is capable of learning synergistic grasping and throwing policies for arbitrary objects in real settings. An after-the-fact analysis of what the deep network learns, shows that the deep features internal to TossingBot effectively use visual appearance to cluster objects based on geometric and physical attributes -- without any explicit supervision other than the goal to throw with accuracy. Qualitative results (videos) are available at \href{http://tossingbot.cs.princeton.edu/}{http://tossingbot.cs.princeton.edu}

\section{Related Work}
\label{sec:related-work}


\mypara{Analytical models for throwing.} Many previous systems built for throwing \cite{mason1993dynamic,gai2013motion,mori20091,senoo2008high,taylor2019optimal} rely on handcrafting or approximating dynamics based on frictional rigid body mechanics, and then optimizing control parameters to execute a throw such that the projectile (typically a ball) lands at a target location. However, as highlighted in \citet{mason1993dynamic}, accurately modeling throwing dynamics is challenging. It requires knowledge of physical properties that are difficult to estimate (\eg aerodynamics, inertia, coefficients of restitution, friction, shape, mass distribution etc.) for both objects and manipulators. As a result, these model-based systems often observe limited throwing accuracy (\eg 40\% success rate in \cite{senoo2008high}), and have difficulty generalizing to changing dynamics over time (\eg deteriorating friction on gripper finger contact surfaces from repeated throwing). In our work, we leverage deep learning and self-supervision to compensate for the dynamics that are not explicitly accounted for in contact/ballistic models, and we train our policies online via trial and error so that they can adapt to new situations on the fly (\eg new object and manipulator dynamics).
\mypara{Learning models for throwing.} More recently, learning-based systems for robotic throwing \cite{aboaf1988task,hu2010ball,kober2011reinforcement,ghadirzadeh2017deep} have also been proposed, which ignore low-level dynamics and directly optimize for task-level success signals (\eg did the projectile land on the target?). These methods have demonstrated better accuracy than those which solely rely only on analytical models, but have two primary drawbacks: 1) limited generalization to new object types (beyond balls, blocks, or darts), and 2) limited pre-throw conditions (\eg human operators are required to manually reset objects and manipulators to match a prescribed initial state before every throw), which makes training from trial and error costly. Both drawbacks prevent their use in real unstructured settings. 


In contrast to prior work, we make no assumptions on the physical properties of thrown objects, nor do we assume that the objects are at a fixed pose in the gripper before each throw. Instead, we propose an object-agnostic pick-and-throw formulation that jointly learns to acquire its own pre-throw conditions (via grasping) while learning throwing control parameters that compensate for varying object properties and dynamics. The system learns through self-supervised trial and error, and resets it own training so that human intervention is kept at a minimum.

\begin{figure}[t]
  \centering
  \includegraphics[width=\linewidth]{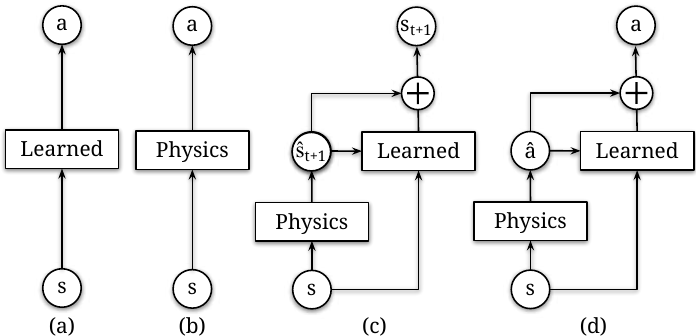}
  \caption{Learning residual models and policies: (a) analytical solutions that determine action $a$ from state $s$; (b) data-driven policies that learn the direct mapping from states to actions; (c) hybrid models that combine analytical models with learning to predict future states $s_{t+1}$; (d) hybrid policies (like ours) that combine analytical solutions with learning to determine action $a$.} 
  \label{fig:model-variants}
  \vspace{-1em}
\end{figure}

\mypara{Learning residual models and policies.} Our approach to data-efficient learning, \textit{Residual Physics}, falls under a broader category of hybrid controllers \cite{abbeel2006using,higuera2017adapting,pastor2013learning} that leverage both 1) analytical models to provide initial estimates of control parameters, and 2) learned residuals on top of those estimates to compensate for unknown dynamics (see Fig. \ref{fig:model-variants}d).
In contrast to prior work on learning residuals on predictions of future states for model-based control \cite{ajay2018augmenting,kloss2017combining} or data-augmented models \cite{fazeli2017learning,jiang2018data,xu2019densephysnet}, we instead directly learn the residuals on control parameters (\ie action space) with deep networks. This approach provides a wider range of data-driven corrections that can compensate for noisy observations as well as dynamics that are not explicitly modeled. These benefits are also observed in concurrent work on residual reinforcement learning \cite{johannink2018residual,silver2018residual} in block-assembly and object manipulation tasks. 


\begin{figure*}[t]
\centering
  \includegraphics[width=\textwidth]{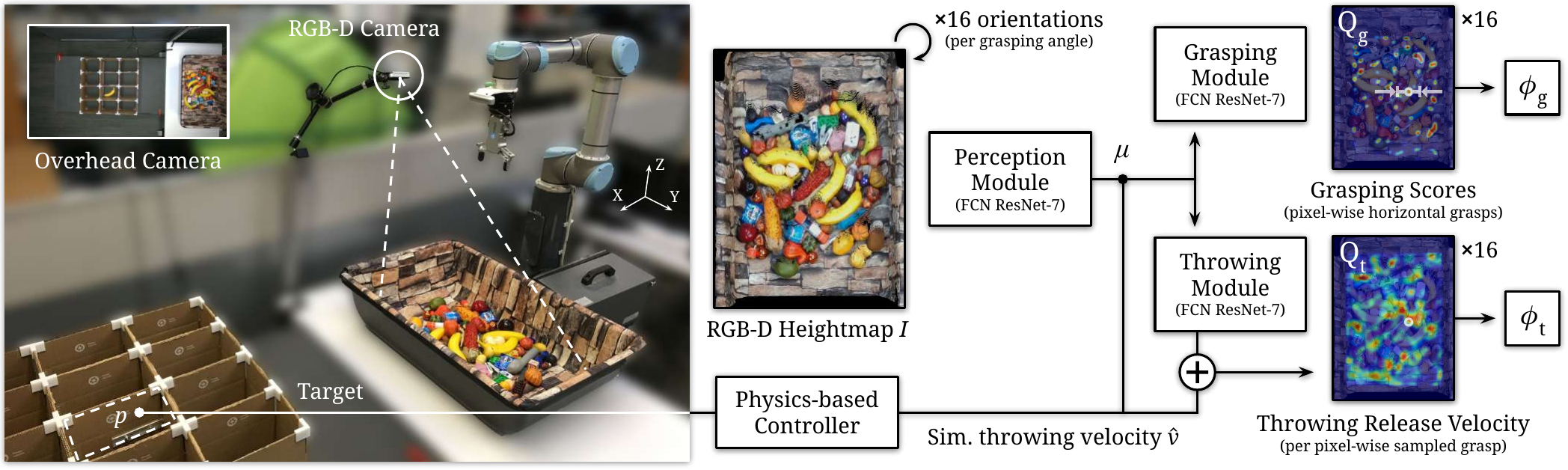}
  \caption{\textbf{Overview.} An RGB-D heightmap of the scene is fed into a perception module to compute spatial features $\mu$. In parallel, target location $p$ is fed into a physics-based controller to provide an initial estimate of throwing release velocity $\hat{v}$, which is concatenated with $\mu$ then fed into grasping and throwing modules. Grasping module predicts probability of grasp success for a dense pixel-wise sampling of horizontal grasps, while throwing module outputs dense prediction of residuals (per sampled grasp), which are added to $\hat{v}$ to get final predictions of throwing release velocities. We rotate input heightmaps by 16 orientations to account for 16 grasping angles. Robot executes the grasp with the highest score, followed by a throw using its corresponding predicted velocity.}
  \label{fig:method-overview}
  \vspace{-1em}
\end{figure*}

\section{Method Overview}
\label{sec:method}

TossingBot consists of a neural network $f(I,p)$ that takes as input a visual observation $I$ of objects in a bin and the 3D position of a target landing location $p$, and outputs a prediction of parameters $\phi_g$ and $\phi_t$ used by two motion primitives for grasping and throwing respectively (see Fig.~\ref{fig:method-overview}). 
The learning objective is to optimize the predictions of parameters $\phi_g$ and $\phi_t$ such that executing the grasping primitive using $\phi_g$ followed by the throwing primitive using $\phi_t$ results in an object (observed in $I$) landing on $p$ at each time-step.

The network $f$ consists of three parts: 1) a perception module that accepts visual input $I$ and outputs a spatial feature representation $\mu$; this is shared as input into 2) a grasping module that predicts $\phi_g$ and 3) a throwing module that predicts $\phi_t$. $f$ is trained end-to-end through self-supervision from trial and error by tracking the ground truth landing positions of thrown objects.  The following subsections provide an overview of these three modules, while the next two sections delve into details of the most novel aspects of the system.




\subsection{Perception Module: Learning Visual Representations}
\label{sec:visual-representation}

We represent the visual input $I$ as an RGB-D heightmap image of the workspace (\ie a bin of objects). To compute this heightmap, we capture RGB-D images from a fixed-mount camera, project the data onto a 3D point cloud, and orthographically back-project upwards in the gravity direction to construct a heightmap image representation with both color (RGB) and height-from-bottom (D) channels. The RGB and D channels are normalized (mean-subtracted and divided by standard deviation from a pre-recorded dataset of 100 images) so that learned convolutional filters can be shared across the two modalities. 

The edges of the heightmaps are defined with respect to the boundaries of the robot's picking workspace. In our experiments, this area covers a $0.9\times0.7$m tabletop surface, on top of which we place a bin of objects. Our heightmaps have a pixel resolution of $180\times140$, hence each pixel $i\in{I}$ represents a $5\times5$mm vertical column of 3D space in the robot's workspace. Using its height-from-bottom value, each pixel thereby corresponds to a unique 3D location in the robot's workspace. The input $I$ is fed into the perception network, a 7-layer fully convolutional residual network \cite{badrinarayanan2015segnet,he2016deep,long2015fully} (interleaved with 2 layers of spatial $2\times2$ max-pooling), which outputs a spatial feature representation $\mu$ of size $45\times35\times512$ that is then fed into the grasping and throwing modules. 


\subsection{Grasping Module: Learning Parallel-jaw Grasps}
\label{sec:grasping}

The grasping module consists of a grasping network that predicts the probability of grasping success for a predefined grasping primitive across a dense pixel-wise sampling of end effector locations and orientations in $I$. 

\mypara{Grasping primitive.} The grasping primitive takes as input parameters $\phi_{g}=(x,\theta)$ and executes a top-down parallel-jaw grasp centered at a 3D location $x=(x_x,x_y,x_z)$ oriented $\theta^\circ$ around the gravity direction. During execution, the open gripper approaches $x$ along the gravity direction until the 3D position of the middle point between the gripper fingertips meets $x$, at which point the gripper closes, and lifts upwards $10$cm. This primitive is open-loop, with robot arm motion planning executed using a stable, collision-free IK solver~\cite{diankov_thesis}.

\mypara{Grasping network.} The grasping network is a 7-layer fully convolutional residual network \cite{badrinarayanan2015segnet,he2016deep,long2015fully} (interleaved with 2 layers of spatial bilinear $2\times$ upsampling). This accepts the visual feature representation $\mu$ as input, and outputs a probability map $Q_g$ with the same image size and resolution as that of the input heightmap $I$. 
Each value of a pixel $q_i\in{Q_g}$ represents the predicted probability of grasping success (\ie grasping affordance) when executing a top-down parallel-jaw grasp centered at the 3D location of $i\in{I}$ with the gripper oriented horizontally with respect to the heightmap ${I}$.

As in \cite{zeng2018robotic,zeng2018learning,zeng2019thesis,zakka2020form2fit}, we account for different grasping angles by rotating the input heightmap by 16 orientations (multiples of $22.5^\circ$) before feeding into the network. Rotations are done with nearest-neighbor sampling to avoid blurring artifacts, as well as 0-padding to maintain fixed image sizes.
The pixel with the highest predicted probability among all 16 maps determines the parameters $\phi_{g}=(x,\theta)$ for the grasping primitive to be executed: the 3D location of a pixel determines the grasping position $x$, and the orientation of the heightmap determines grasping angle $\theta$. This visual state and action representation has been shown to provide sample efficiency when used in conjunction with fully-convolutional action-value functions for grasping and pushing~\cite{zeng2018learning,zeng2018robotic}. Each pixel-wise prediction shares convolutional features for all grasping locations and orientations (\ie translation and rotation equivariance). 





\subsection{Throwing Module: Learning Throwing Velocities}
\label{sec:throwing}



The goal of the throwing module is to predict the release position and velocity of a predefined throwing primitive for each possible grasp (over the dense pixel-wise sampling of end effector locations and orientations in $I$).

\mypara{Throwing primitive.} The throwing primitive takes as input parameters $\phi_{t}=(r,v)$ and executes an end effector trajectory such that the mid-point between the gripper fingertips reaches a desired release position $r=(r_x,r_y,r_z)$ and velocity $v=(v_x,v_y,v_z)$, at which point the gripper opens and releases the object. During execution, the robot arm curls inwards while grasping onto an object, then uncurls outward at high speed, releasing the object at the desired position and velocity. Throughout this motion, the gripper is oriented such that the axis between the fingertips is orthogonal to the plane of the intended aerial trajectory. In our system, the direction of curling/uncurling aligns with $(v_x,v_y)$. Fig. \ref{fig:grasping-affects-tossing} visualizes this motion primitive and its end effector trajectory. The throwing primitive is executed after each successful grasp attempt (checked by thresholding the distance between fingertips).

\mypara{Planning the release position.} In most real-world settings, only a handful of release positions are physically accessible by the robot for throwing. So for simplicity in our system, we directly constrain the release position $r$ from the given target landing location $p$ using two assumptions: 1) the aerial trajectory of a projectile is linear on the xy-horizontal-plane and in the same direction as $v_{x,y}=(v_x,v_y)$. In other words, we assume that the forces of aerodynamic drag \textit{orthogonal} to $v_{x,y}$ are negligible. This is not to be confused with the primary forces of drag that exist in \textit{parallel} to $v_{x,y}$, which our system will learn to compensate. We also assume 2) that the release distance $\sqrt{r_x^2+r_y^2}$ is at a fixed value $c_d$ from the robot base origin, and that $r_z$ is at a fixed constant height $c_h$. Formally, these constraints can be written as: 
$(r_{x,y}-p_{t_{x,y}})\times{v_{x,y}}=0$
and $\sqrt{r_x^2+r_y^2} = c_d$ and $r_z=c_h$. 
In our experiments, we select constant values of $c_h$ and $c_d$ such that all release positions are accessible by the robot: $c_h=0.04$m and $c_d=0.7$m in simulation, and $c_h=0.02$m and $c_d=0.76$m in real settings. 

\mypara{Planning the release velocity.} Given a target landing location $p$ and release position $r$, there could be multiple solutions of the release velocity $v$ for which the object lands on $p$. To remove this ambiguity, we further constrain the direction of $v$ to be angled $45^\circ$ upwards in the direction of $p$. Formally, this constraint can be defined as $\|v_{x,y}\|=v_z$.
Under all the aforementioned constraints, the only unknown variable for throwing is $\|v_{x,y}\|$, which represents the magnitude of the final release velocity. 
Specifically, assuming a fixed throwing release height $r_z$, fixed release distance $c_d$ from robot base origin, and release velocity direction angled $45^\circ$ upwards: for any given target landing location $p=(p_x,p_y,p_z)$, we can derive a release position $r$ and release velocity magnitude $\|v\|$ that achieves the target landing location $p$ assuming equations of linear projectile motion: 
\begin{equation}
  \begin{split}
  \theta &= \arctan{(\frac{p_y}{p_x})}\\
  r_x &= c_d\sin(\theta)\\
  r_y &= c_d\cos(\theta)
  \end{split}
\end{equation}

\begin{equation}
  \|v\| = \sqrt{\frac{a(p_x^2+p_y^2)}{(r_z-p_z-\sqrt{p_x^2+p_y^2})}}
\end{equation}
where $a$ is acceleration from gravity.

These equations are valid for any given target landing location $p$, as long as both $\|v\|$ and $r$ are within robot physical limits.  Hence assuming no aerial obstacles, varying only the velocity magnitude $\|v\|$ is sufficient to cover the space of all possible projectile landing locations.  
In the following section, we describe how the throwing module predicts $\|v_{x,y}\|$.

\section{Learning Residual Physics for Throwing}
\label{sec:residual-physics}

A key aspect of TossingBot's throwing module is that it learns to predict a residual $\delta$ on top of the estimated release velocity $\|\hat{v}_{x,y}\|$ from a physics-based controller (\ie ballistic equations of projectile motion), then uses the superposition of the two predictions to compute a final release velocity $\|v_{x,y}\|=\|\hat{v}_{x,y}\|+\delta$ for the throwing primitive. Conceptually, this enables our models to leverage the advantages of physics-based controllers (\eg generalization via analytical models), while still maintaining the capacity (via data-driven residual $\delta$) to account for aerodynamic drag and offsets to the real-world projectile velocity (conditioned on the grasp), which are otherwise not analytically modeled. Our experiments in Sec. \ref{sec:evaluation} show that this approach, a.k.a. \textit{Residual Physics}, even when using a simple ballistics model, yields significant improvements in both accuracy and generalization of throwing arbitrary objects compared to baseline alternatives: \eg using only the physics-based controller (Fig. \ref{fig:model-variants}a), or a fully data-driven training of $f$ to regress $\|v_{x,y}\|$ (Fig. \ref{fig:model-variants}b).

\mypara{Physics-based controller.} The physics-based controller uses the standard equations of linear projectile motion, by assuming a grasp on the center of mass of the object, to analytically solve back for the release velocity $\hat{v}$ given the target landing location $p$ and release position $r$ of the throwing primitive:
$p=r+\hat{v}t+\frac{1}{2}at^{2}$.
\noindent This controller assumes that the object is a point particle and that the aerial trajectory of the projectile moves along a ballistic path affected only by gravity, which imparts a downward acceleration $a_z=-9.8$m/s$^{2}$. 

We also provide the estimated physics-based release velocity $\hat{v}$ as input into both the grasping and throwing networks by concatenating the visual feature representation $\mu$ with a $k$-channel image ($k=128$) where each pixel holds the value of $\hat{v}$, repeated across channels. Providing $\hat{v}$ as input enables our grasping and throwing predictions to be conditioned on $\hat{v}$ -- \ie larger values of $\hat{v}$ for farther target locations can lead to a different set of effective grasps.

This physics-based controller has several advantages in that it provides a closed-form solution, generalizes well to new landing locations $p$, and serves as a consistent approximation for $v$. However, it also relies on several assumptions that do not generally hold. First, it assumes that the effects of aerodynamic drag are negligible. However, as we show in our experiments in Fig. \ref{fig:grasping-affects-tossing}, the aerial trajectory for lightweight objects like ping pong balls can be substantially influenced by drag. Second, it assumes that the gripper release velocity $v$ directly determines the velocity of the projectile. This is not true since the object is often not grasped at the center of mass, nor is the object completely immobilized by the grasp prior to release. For example, as illustrated in Fig. \ref{fig:grasping-affects-tossing}, a screwdriver picked up by the shaft can be flung forward with a significantly higher velocity than the gripper release velocity due to centripetal forces, resulting in a farther aerial trajectory.

\mypara{Residual physics-based controller.} To compensate for the shortcomings of the physics-based controller, the throwing module includes a throwing network that predicts a residual $\delta$ on top of the estimated release velocity $\|\hat{v}_{x,y}\|$ for each possible grasp. The throwing network is a 7-layer fully convolutional residual network \cite{he2016deep} interleaved with 2 layers of spatial bilinear $2\times$ upsampling that accepts the visual feature representation $\mu$ as input, and outputs an image $Q_t$ with the same size and resolution as that of the input heightmap $I$. ${Q_t}$ has a pixel-wise one-to-one spatial correspondence with $I$, thus each pixel in $Q_t$ also corresponds one-to-one with the pixel-wise probability predictions of grasping success $q_i\in{Q_g}$ (for all possible grasps using rotating input $I$). Each pixel in $Q_t$ holds a prediction of the residual value $\delta_i$ added on top of the estimated release velocity $\|\hat{v}_{x,y}\|$ from a physics-based controller, to compute the final release velocity $v_i$ of the throwing primitive after executing the grasp at pixel $i$.
The better the prediction of $\delta_i$, the more likely the grasped and thrown object will land on the target location $p$.

\section{Jointly Learning Grasping and Throwing}
\label{sec:learning-objectives}

Our full network $f$ (including the perception, grasping, and residual throwing modules) is trained end-to-end using the following loss function: $\mathcal{L}=\mathcal{L}_g+y_i\mathcal{L}_t$,
where $\mathcal{L}_g$ is the binary cross-entropy error from predictions of grasping success classification:
$$\mathcal{L}_g=-(y_i\log{q_i}+(1-y_i)\log(1-q_i))$$
and $\mathcal{L}_t$ is the continuous Huber loss from regressing the residual throwing velocity $\delta_i$:
$$\mathcal{L}_t=\left\{\begin{array}{ll} \frac{1}{2}(\delta_i-\bar{\delta}_i)^2, \mathrm{for}\:|\delta_i-\bar{\delta}_i| < 1,\\
                  |\delta_i-\bar{\delta}_i| - \frac{1}{2},\:\mathrm{otherwise.}\end{array}\right.$$
\noindent where $y_i$ is the binary ground truth grasp success label and $\bar{\delta}_i$ is the ground truth residual label. We use an Huber loss \cite{girshick2015fast} instead of an MSE loss for regression since we find that it is less sensitive to inaccurate outlier labels. We pass gradients only through the single pixel $i$ on which the grasping primitive was executed. All other pixels backpropagate with 0 loss. 

We train our network $f$ by stochastic gradient descent with momentum, using fixed learning rates of $10^{-4}$, momentum of 0.9, and weight decay $2^{-5}$. Our models are trained from scratch (\ie random Xavier initialization \cite{glorot2010understanding}, without any image-based pretraining \cite{yen2020learning}) in PyTorch with an NVIDIA Titan X on an Intel Xeon CPU E5-2699 v3 clocked at 2.30GHz.
We train with prioritized experience replay \cite{schaul2015prioritized} using stochastic rank-based prioritization, approximated with a power-law distribution. Our exploration strategy is $\epsilon$-greedy, with $\epsilon$ initialized at 0.5 then annealed over training to 0.1. Specifically, when executing a grasp, the robot has an $\epsilon$ chance to sample a random grasp within the robot's workspace for picking; likewise when executing a throw, the robot has an $\epsilon$ chance to explore a random positive release velocity. 


\mypara{Network architecture details.} Our network architecture consists of the following layers for each module:
\begin{itemize}
\item Perception: C(3,64)-MP-RB(128)-MP-RB(256)-RB(512)
\item Grasping: RB(256)-RB(128)-UP-RB(64)-UP-C(1,2)
\item Throwing: RB(256)-RB(128)-UP-RB(64)-UP-C(1,1)
\end{itemize}
where C(k,c) denotes a convolutional layer with k$\times$k filters and c channels, RB(c) denotes a residual block \cite{he2016deep} with two convolutional layers using 3x3 filters and c channels, MP denotes a 3$\times$3 max pooling layer with stride = 2, and UP denotes a bilinear 2$\times$ upsampling layer.

\mypara{Training via self-supervision.} We obtain our ground truth training labels $y_i$ and $\bar{\delta}_i$ through trial and error. At each training step, the robot captures RGB-D images to construct visual input $I$, performs a forward pass of $f(I,p)$ to make a prediction of primitive parameters $\phi_g$ and $\phi_t$, executes the grasping primitive using $\phi_g$, then executes the throwing primitive using $\phi_t$. 
We obtain ground truth grasp success labels $y_i$ by one of two ways:
\begin{itemize}[noitemsep]
\item[1.] Success after grasping, by checking the distance between gripper fingertips after the grasping primitive.
\item[2.] Success after throwing, by checking the binary signal of whether or not a throw lands in the correct box.
\end{itemize}
As we show in Sec. \ref{sec:evaluation-stable-grasps}, supervising grasps by the accuracy of throws eventually leads to more stable grasps and better overall throwing performance. The grasping policy learns to favor grasps that lead to successful throws, which is a stronger requirement than simple grasp success. 
After each throw, we approximate the object's actual landing location $\hat{p}$ using a calibrated overhead RGB-D camera to detect which bin it landed in (\ie detect visual changes in the landing zone before and after the throw). 
Regardless of where the object lands, its actual landing location $\hat{p}$ and the executed release velocity $v$ is recorded and saved to the experience replay buffer as a training sample, with which we obtain the ground truth residual label $\bar{\delta}_{i}=\|v_{x,y}\|-\|\hat{v}_{x,y}\|_{\hat{p}}$.

In experiments in Sec. \ref{sec:evaluation}, we train our models by self-supervision with the same procedure: $n$ objects are randomly dropped into the $0.9\times0.7$m workspace in front of the robot. The robot performs data collection until the workspace is void of objects, at which point $n$ objects are again randomly dropped into the workspace. In simulation $n=12$, while in real-world experiments $n=80$+. In our real-world setup, the landing zone (on which target boxes are positioned) is slightly tilted at a $15^\circ$ angle adjacent to the bin to simplify resetting. When the workspace is void of objects, the robot lifts the bottomless boxes such that the objects slide back into the bin. In this way, human intervention is kept at a minimum during the training process. Videos of this reset process are available at \href{http://tossingbot.cs.princeton.edu/}{http://tossingbot.cs.princeton.edu}

\section{Evaluation}
\label{sec:evaluation}

We execute a series of experiments in simulated and real settings to evaluate the learned grasping and throwing policies. The goal of the experiments are four-fold:
1) to evaluate the overall accuracy and efficiency of our pick-and-throw system on arbitrary objects, 
2) to test its generalization to new objects and target locations unseen during training, 
3) to investigate how learned grasps can improve the accuracy of subsequent throws,  and
4) to compare our proposed method based on \textit{Residual Physics} to other baseline alternatives.  

\mypara{Evaluation metrics} are 1) grasping success: the \% rate which an object remains in the gripper after executing the grasping primitive (by measuring distance between fingertips), and 2) throwing success: the \% rate which a thrown object lands in the intended target box (tracked by an overhead camera).
 
\begin{figure}[t]
\centering
  \includegraphics[width=\linewidth]{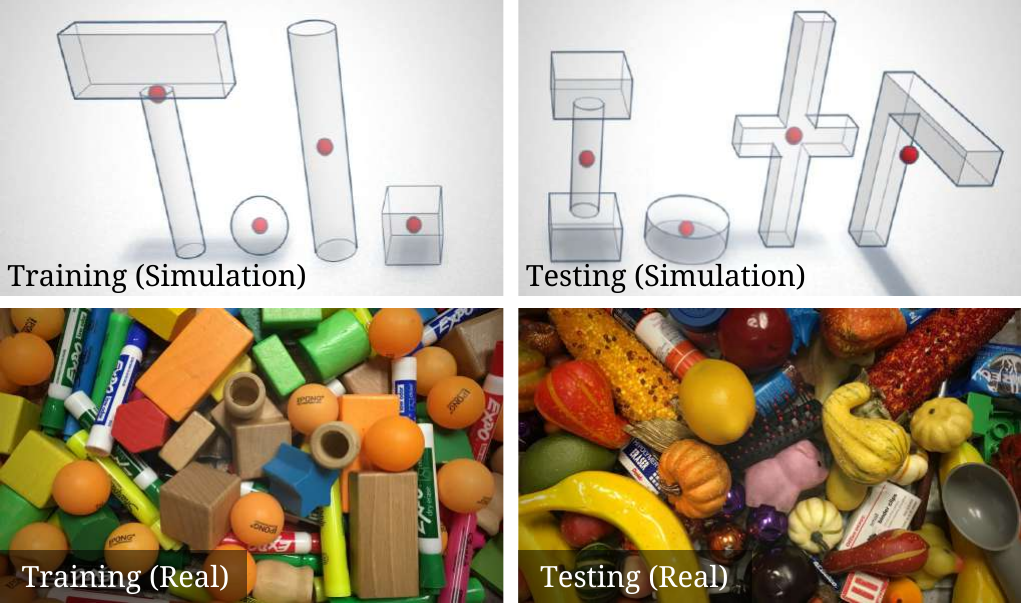}
  \caption{\textbf{Objects} used in simulated (top) and real (bottom) experiments, split by seen objects (left) and unseen objects (right). The center of mass of each simulated object is indicated with a red sphere (for illustration).}
  \label{fig:objects}
  \vspace{-1em}
\end{figure}

\subsection{Experimental Setup} 

We evaluate each policy on its ability to grasp and throw various objects into 12 boxes located outside a UR5 robot arm's maximum reach range (as shown in Fig.~\ref{fig:teaser}). Specifically, the task is to pick objects from a cluttered bin and stow them uniformly into the boxes such that all boxes have the same number of objects, regardless of object type. Since boxes are \textit{outside} the robot's reach range, throwing is necessary to succeed in the task. Each box is $20$cm tall with a $25\times15$cm opening. The middle of the top opening of each box is used as the input target landing position $p$ to the formulation $f(I,p)$.

\begin{figure}[t]
  \centering
  \includegraphics[width=0.99\linewidth]{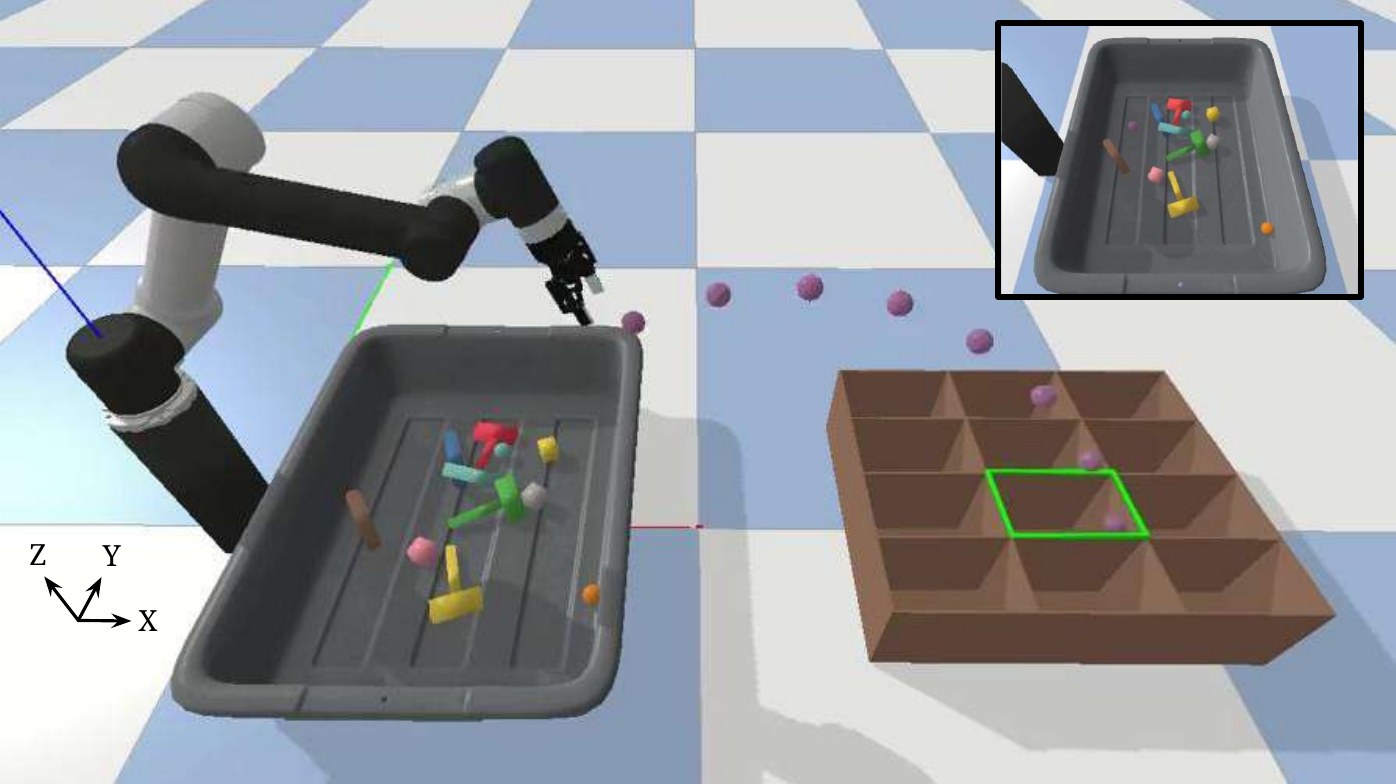}
  \caption{\textbf{Simulation environment} in PyBullet \cite{coumans2018}. This snapshot illustrates the aerial motion trajectory of a purple ball being thrown into the target landing box highlighted in green. The top right image depicts the view captured from the simulated RGB-D camera before the ball was grasped and thrown.}
  \label{fig:simulation-environment}
  \vspace{-1em}
\end{figure}

\mypara{Simulation setup}. The simulation environment (shown in Fig. \ref{fig:simulation-environment}) is built using PyBullet \cite{coumans2018}. We use 8 different objects: 4 seen during training and 4 unseen for testing. Training objects are chosen in order of increasing difficulty: $4$cm-diameter ball, $4\times4\times4$cm cube, $3$cm-diameter $16$cm-long rod, and a $16$cm-long hammer (union of $2$cm-diameter $12$cm-long rod with $10\times4\times2.5$cm block). Throwing difficulty is determined by how much an object's projectile trajectory changes depending on its initial grasp and center of mass (CoM). For example, the trajectory of the ball is mostly agnostic to grasp location and orientation, while both rod (CoM in middle) and hammer objects (CoM between handle and shaft) can have drastically different projectile trajectories depending on the grasping point. Objects are illustrated in Fig. \ref{fig:objects} -- their CoMs indicated with a red sphere. Multiple copies of each object (12 in total) are randomly colored and dropped into the bin during training and testing.
 
Although simulation provides a consistent and controlled environment for fair ablative analyses, the simulated environment does not model aerodynamics and only approximates frictional interactions. As a result, performance in simulation does not necessarily reflect performance in the real world. Therefore we also provide quantitative experiments on a real system.

\mypara{Real-world setup}. 
We use a UR5 arm with an RG2 gripper to pick and throw a collection of 80+ different toy blocks, fake fruit, decorative items, and office objects (see Fig. \ref{fig:objects}). For perception data, we capture $640\times480$ RGB-D images using a calibrated Intel RealSense D415 statically mounted overlooking the bin of objects from the side. The camera is localized with respect to the robot base using an automatic calibration procedure from \cite{zeng2018learning}. A second RealSense D415 is mounted above the boxes looking downwards to track landing locations of thrown objects by measuring changes between images captured before and after executed throws. 
  

\subsection{Baseline Methods}
\label{sec:evaluation-baselines}

\mypara{Residual-physics} denotes our approach described in Sec. \ref{sec:method}. Since there are no comparable available algorithms that can learn joint grasping and throwing policies, we compare our approach to three baselines based on variations of the proposed method:

\mypara{Regression} is a variant of our approach where the throwing network is trained to directly regress the final release velocity $v$, instead of the residual $\delta$. Specifically, each pixel in the output $Q_t$ of the throwing network holds a prediction of the final release velocity $\|v_{x,y}\|$ for the throwing primitive. The physics-based controller is removed from this baseline, but in order to ensure a fair comparison, we concatenate the visual features $\mu$ with the xy-distance $d$ between the target landing location and release position (\ie $d=\|r_{x,y}-p_{t_{x,y}}\|$) before feeding into the grasping and throwing networks. Conceptually, this variant of our approach is forced to learn physics from scratch instead of bootstrapping on physics-based control.

\mypara{Physics-only} is also a variant of our approach where the throwing network is removed and completely replaced by velocity predictions made by the physics-based controller. In other words, this variant only learns grasping and uses physics for throwing (without learning a residual). 

\mypara{Regression-pretrained-on-physics} is a version of \textbf{\textit{Regression}} that is pre-trained on release velocity predictions $\hat{v}$ made by the physics-based controller described in Sec. \ref{sec:throwing}. The shorthand name for this method is \textbf{\textit{Regression-PoP}}.


\subsection{Baseline Comparisons}
In simulated and real settings, we train our models via trial and error for 15,000 steps, then test each model for 1,000 steps and report their average grasping and throwing success rates. 

\mypara{Simulation results} are reported in Table \ref{table:main-sim-throw} and \ref{table:main-sim-grasp}. Each column of the table represents a different set of test objects e.g., ``Hammers'' is a set of $n$ hammers, ``Seen'' is a mixed set of objects seen during training, ``Unseen'' is a mixed set of objects not seen during training.

\begin{table}[h]
  \centering
  \setlength{\tabcolsep}{3.5 pt}
  \caption{Throwing Performance in Simulation (Mean \%)}
  \vspace{-0.5em}
  \begin{tabular}{lcccccc}
  \toprule
  Method & Balls & Cubes & Rods & Hammers & Seen & Unseen \\ 
  \midrule
  Regression       & 70.9 & 48.8 & 37.5 & 32.8 & 41.8 & 28.4 \\    
  Regression-PoP   & 96.1 & 73.5 & 52.8 & 47.8 & 56.2 & 35.0 \\   
  Physics-only     & 98.6 & 83.5 & 77.2 & 70.4 & 82.6 & 50.0 \\  
  Residual-physics & \bf{99.6} & \bf{86.3} & \bf{86.4} & \bf{81.2} & \bf{88.6} & \bf{66.5} \\  
  \bottomrule
  \label{table:main-sim-throw}
  \end{tabular}
  
  \centering
  \setlength{\tabcolsep}{3.5 pt}
  \caption{Grasping Performance in Simulation (Mean \%)}
  \vspace{-0.5em}
  \begin{tabular}{lcccccc}
  \toprule
  Method & Balls & Cubes & Rods & Hammers & Seen & Unseen \\
  \midrule
  Regression       & 99.4 & 99.2 & 89.0 & 87.8 & 95.6 & 69.4 \\  
  Regression-PoP   & 99.2 & 98.0 & 89.8 & 87.0 & 96.4 & 70.6 \\   
  Physics-only     & 99.4 & 99.2 & 87.6 & 85.2 & 96.6 & 64.0 \\    
  Residual-physics & 98.8 & 99.2 & 89.2 & 84.8 & 96.0 & 74.6 \\  
  \bottomrule
  \label{table:main-sim-grasp}
  \end{tabular}
  \vspace{-1em}
\end{table}

The throwing results in Table \ref{table:main-sim-throw} indicate that learning residuals (Residual-physics) on top of a physics-based controller provides the most accurate throws. Physics-only performs competitively in simulation, where the environment is void of aerodynamics and unstable contact dynamics, but falls short of performance in comparison to Residual-physics -- particularly for difficult objects like rods or hammers of which the grasping offsets from CoM can significantly change projectile trajectories. We also observe that regression pre-trained on physics (Regression-PoP) always consistently outperforms regression alone. On the other hand, the results in Table \ref{table:main-sim-grasp} show that grasping performance remains roughly the same across all methods. All policies experience moderately lower grasping and throwing success rates for unseen testing objects. 

\begin{figure}[t]
  \centering
  \vspace{-1.5em}
  \includegraphics[width=\linewidth]{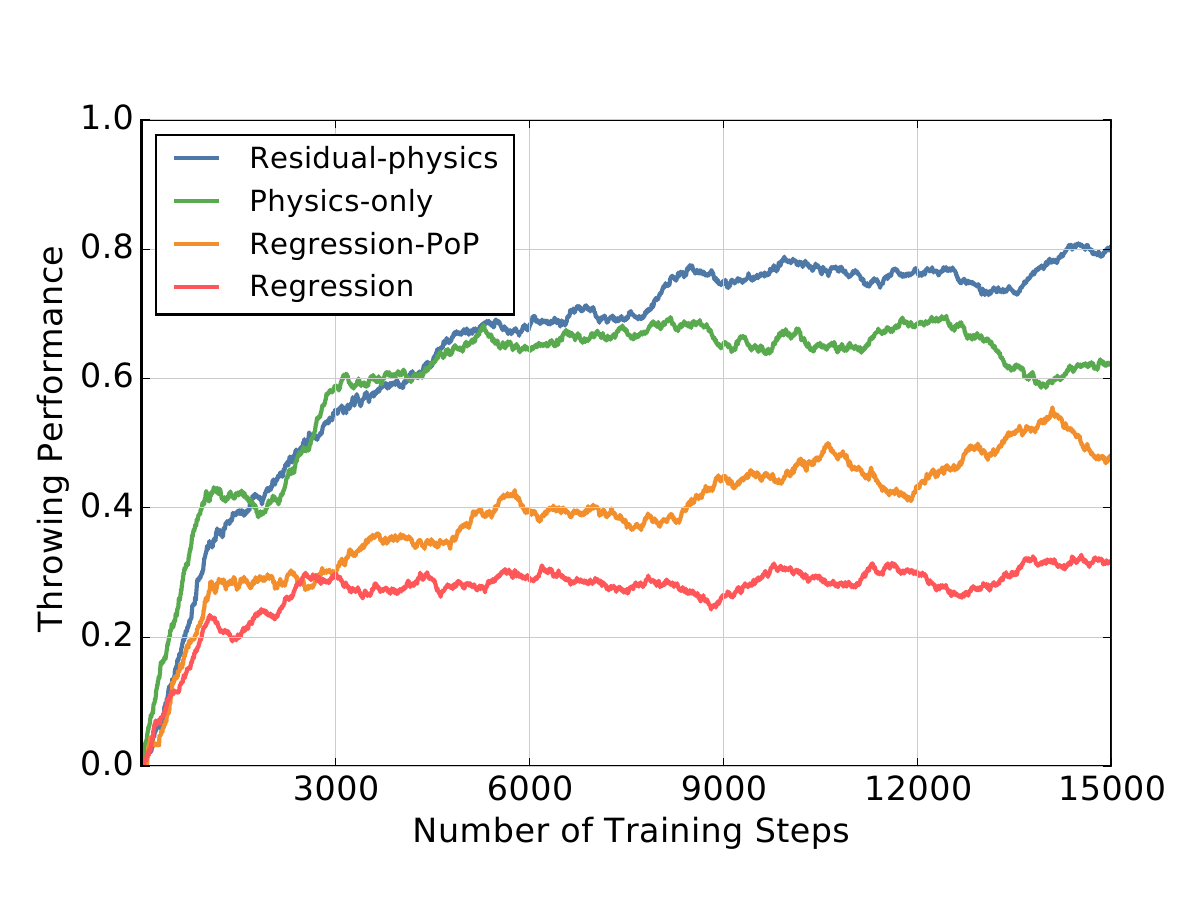}
  \vspace{-2.5em}
  \caption{Our method (Residual-physics) outperforms baseline alternatives in terms of throwing success rates in simulation on the Hammers object set.
  }
  \label{fig:main-sim-throw-hammer}
  \vspace{-1em}
\end{figure}

Fig. \ref{fig:main-sim-throw-hammer} plots the average throwing performance of all baseline methods over training steps on the hardest seen object set: hammers. Throwing performance is measured by throwing success rates over the last $j=\textrm{1,000}$ attempts. Numbers reported at earlier training steps (\ie iteration $i<j$) in Fig. \ref{fig:main-sim-throw-hammer} are weighted by $\frac{i}{j}$. The plot shows that as soon as the performance of Physics-only begins to asymptote, Residual-physics starts to outperform Physics-only by learning residual throwing velocities that compensate for grasping offsets from the object CoM.

\mypara{Real-world results} are reported in Table \ref{table:performance-real} on seen and unseen object sets. The results show that Residual-physics continues to provide more accurate throws than the baseline methods. Most notably, in contrast to simulation, Physics-only does not perform as competitively to Residual-physics in the real-world. This is likely because the ballistic model used by Physics-only does not account for the unmodelled and uncertain contact- and aero-dynamics in the real world. Residual-physics can compensate for them in one of two ways: either improving the model (learning good residuals), or avoiding regions of the model that are not predictable (avoiding complex grasps). This allows TossingBot to maintain a throwing accuracy above 80\% for both seen and unseen objects. 

Interestingly, our system also seems to match or even slightly exceed the average performance of an untrained human (\ie with no training time provided beforehand). To measure human throwing performance, we asked 15 willing participants (average height: 174.0\rpm8.3cm, most of whom were engineers) to stand in place of the robot in the real-world setup and then grasp and throw 80 objects from the bin into the target boxes round-robin. Objects came from the collection of unseen test objects used in the robot experiments, and were kept consistent across runs. Participants were asked to pick-and-throw at whichever speed felt most comfortable (\ie we did not compare picking speeds).

Surprisingly, human performance was lower than we  expected. The largest contributor to poor performance was fatigue -- the accuracy of throws deteriorates over time, particularly after around the 20th object regardless of picking speed. The second largest contributor to performance was the physical height of the participant (taller performed better) -- this may be due to differences in throwing distance (measured from grasp release to object landing locations, which is smaller for taller participants with longer arms) and the throwing strategies (taller participants more often preferred overhand throws to underhand ones). Other common throwing strategies included: 1) largely relying tactile feedback to grasp objects in the bin to maintain visual attention on target boxes, 2) grasping objects with one hand and throwing with the other so that the throwing arm can make more repeatable movements, 3) and grouping objects by weight, then correspondingly changing to different grasping and throwing strategies. These additional strategies were interesting, but did not seem to correlate with better performance. Also, most strategies seem designed to overcome human limitations in terms of restricted attention spans, limited viewpoints, limited motor control calibration, or fatigue, which do not hinder robotic systems.


\begin{table}[h]
  \centering
  \caption{Grasping and Throwing Performance in Real (Mean \%)}
  \vspace{-1em}
  \begin{tabular}{lcccc}
  \toprule
  & \multicolumn{2}{c}{Grasping} & \multicolumn{2}{c}{Throwing}\\
  \cmidrule{2-3}
  \cmidrule{4-5}
  Method & Seen & Unseen & Seen & Unseen\\
  \midrule
  Human-baseline   & --   & --   & --   & 80.1\rpm10.8 \\
  Regression-PoP   & 83.4 & 75.6 & 54.2 & 52.0 \\   
  Physics-only     & 85.7 & 76.4 & 61.3 & 58.5 \\    
  Residual-physics & 86.9 & 73.2 & \bf{84.7} & \bf{82.3} \\  
  \bottomrule
  \label{table:performance-real}
  \end{tabular}
  \centering
  \caption{Picking Speed vs State-of-the-Art Systems}
  \vspace{-1em}
  \begin{tabular}{l c c}
  \toprule
  System & Mean Picks Per Hour (MPPH) \\
  \midrule
  Cartman \cite{morrison2018cartman} & 120 \\
  Dex-Net 2.0 \cite{mahler2017dex} & 250 \\
  FC-GQ-CNN \cite{satish2018on} & 296 \\
  Dex-Net 4.0 \cite{mahler2019learning} & 312 \\
  TossingBot (w/ Placing) & 432 \\  
  TossingBot (w/ Throwing) & \bf{514} \\   
  \bottomrule
  \label{table:picking-speed}
  \end{tabular}
  \vspace{-2em}
\end{table}

\subsection{Pick-and-Place Efficiency} 

Throwing enables our system (TossingBot) to achieve picking speeds of 514 mean picks per hour (MPPH), where 1 pick = successful grasp and accurate throw. Specifically, the system performs 608 grasps per hour (measured over two hours), and achieves 84.7\% throwing accuracy, yielding 514 MPPH. In Table \ref{table:picking-speed}, we compare against other state-of-the-art picking systems found in literature: Cartman \cite{morrison2018cartman}, Dex-Net 2.0 \cite{mahler2017dex}, FC-GQ-CNN \cite{satish2018on}, Dex-Net 4.0 \cite{mahler2019learning}, and a variant of TossingBot that places objects into a box 0.8m away from the bin without throwing. This is not a like-for-like comparison, since throwing is only practical for certain types of objects (\eg not eggs) and hardware, and placing is only practical for limited distance ranges. Yet, the results suggest that throwing may be useful to improve the overall MPPH in some applications. 

In addition to throwing, there are 3 other aspects that enable our system's picking speeds: 1) fast algorithmic run-time speeds (220ms for inference), 2) real-time TSDF fusion \cite{curless1996volumetric,newcombe2011kinectfusion,zeng20163dmatch,sajjan2020cleargrasp} of RGB-D data, which enables us to capture and aggregate observed 3D data of the scene simultaneously as the robot moves around within the field-of-view, and 3) online training and inference in parallel to robot actions (described in Algorithm \ref{alg:system-pipeline}.

\begin{algorithm}
\caption{System Pipeline}\label{alg:system-pipeline}
\begin{algorithmic}[1]
  \State Initialize \textit{robot}.
  \State Initialize policy with model $f$.
  \State Initialize replay \textit{buffer}.
  \While{step $i<N$ and not \textit{terminate}}
    \State $I^i$ = \textit{robot}.CaptureState()
    \State $p^i$ = \textit{robot}.SelectTarget()
    \State $\phi^i_g$,$\phi^i_t$ = $f$.Inference($I^i$,$p^i$)
    \While{\textit{robot}.is\_grasping}
      \State $f$.Train(\textit{buffer})
    \EndWhile
    \State $y^{i-1}$ = \textit{robot}.CheckGraspSuccess()
    \State \textit{robot}.ExecuteThrow($\phi^{i-1}_t$,$p^{i-1}$) \Comment{asynchronous}
    \While{\textit{robot}.is\_throwing}
      \State $f$.Train(\textit{buffer})
    \EndWhile
    \State \textit{robot}.ExecuteGrasp($\phi^{i}_g$) \Comment{asynchronous}
    \State $\hat{p\,}^{i-1}$ = \textit{robot}.TrackLanding()
    \State \textit{buffer}.SaveData($I^{i-1}$,$p^{i-1}$,$\phi^{i-1}_g$,$\phi^{i-1}_t$,$y^{i-1}$,$\hat{p\,}^{i-1}$)
    \State $i$ = $i+1$
  \EndWhile
\end{algorithmic}
\end{algorithm}

\subsection{Learning Stable Grasps for Throwing} 
\label{sec:evaluation-stable-grasps}

\begin{figure}[t]
  \centering
  \vspace{-1em}
  \includegraphics[width=\linewidth]{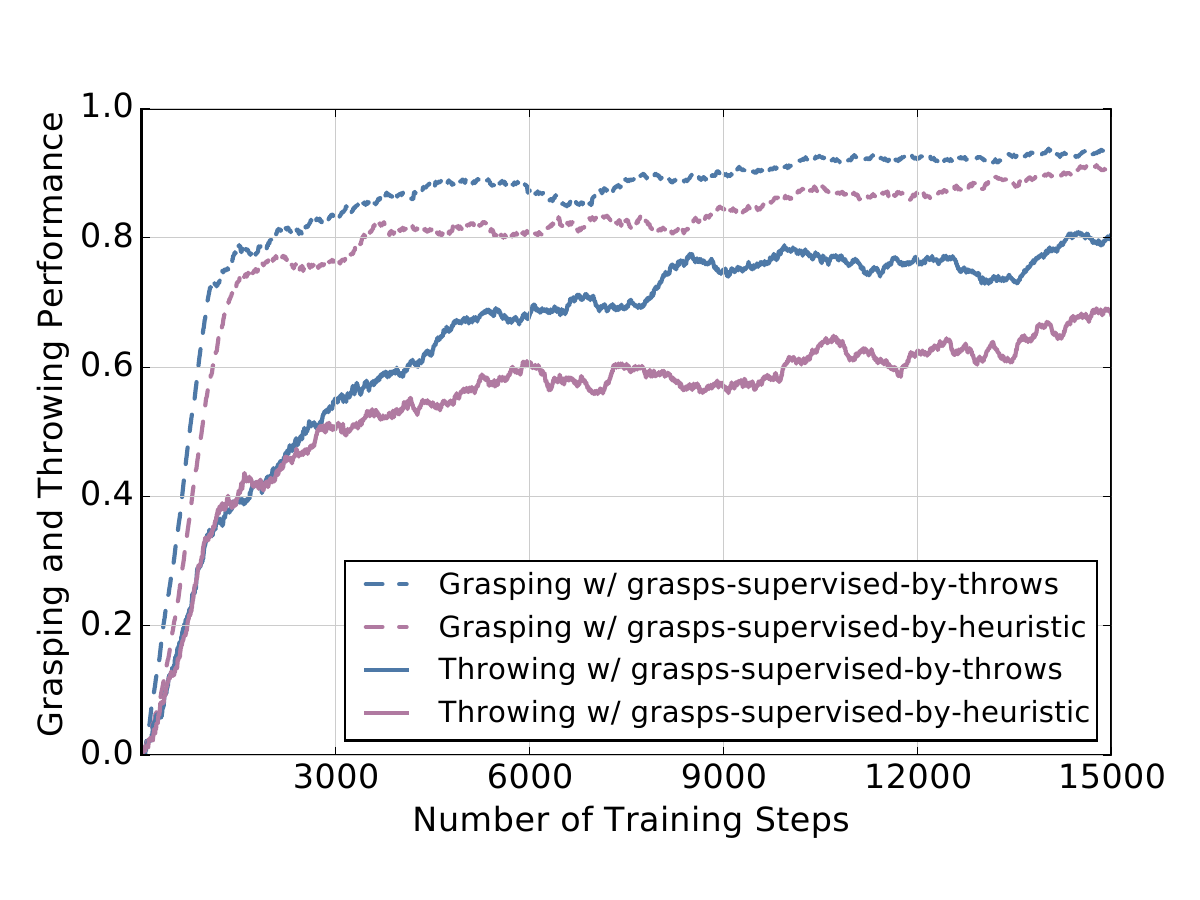}
  \vspace{-2.5em}
  \caption{
  Both grasping and throwing success rates of Residual-physics policies in simulation improve when grasps are supervised by the accuracy of throws (blue), versus when grasps are supervised by a heuristic that checks gripper width (purple).
  }
  \label{fig:graspcheck-sim-throw-hammer}
  \vspace{-1em}
\end{figure}

We next investigate the importance of supervising grasps with the accuracy of throws. To this end, we train two variants of Residual-physics: 1) grasping network supervised by accuracy of throws (\ie grasp success = object landed on target), and 2) grasping network supervised by checking grasp width after grasping primitive (\ie grasp success = object in gripper). We plot their grasping and throwing success rates over training steps in Fig. \ref{fig:graspcheck-sim-throw-hammer} on the hammer object set.

The results indicate that throwing performance significantly improves when grasping is supervised by the accuracy of throws. This not only suggests that the grasping policies are capable of learning to execute the subset of grasps that lead to more predictable throws, but also indirectly that throwing accuracy is strongly influenced by the quality of grasps. Interestingly, the results also show that grasping performance slightly increases when supervised by the accuracy of throws.

\begin{figure}[t]
  \centering
  \includegraphics[width=\linewidth]{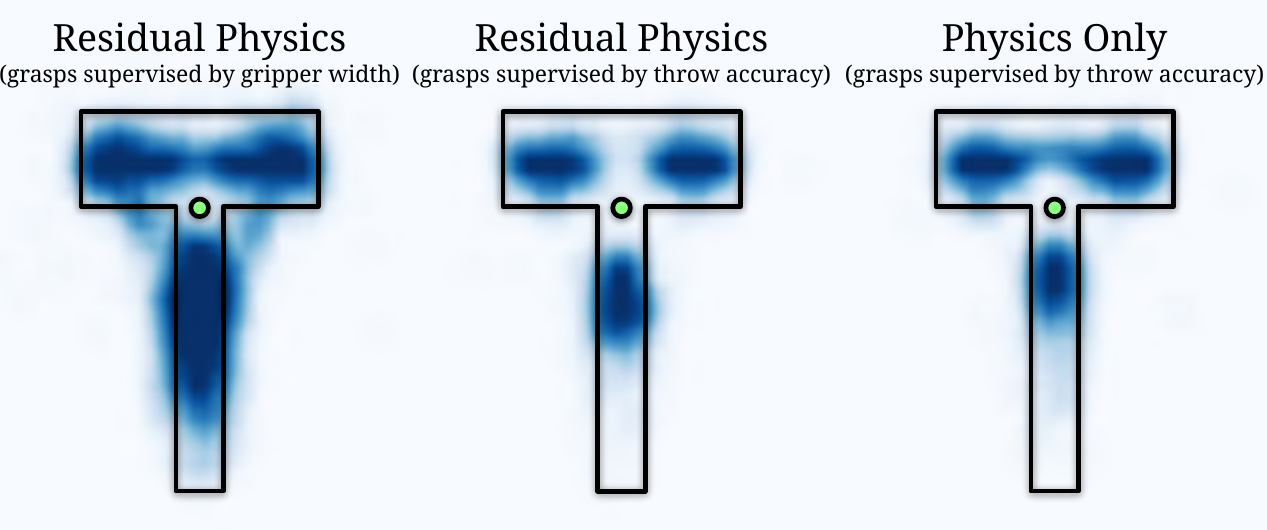}
  \caption{Projected 2D histograms of successful grasping positions on hammers in simulation: show that 1) leveraging accuracy of throws as supervision enables the grasping policy to learn a more restricted but stable set of grasps, while 2) learning throwing in general helps to relax this constraint.}
  \label{fig:grasp-supervision}
  \vspace{-1em}
\end{figure}

We also investigate the quality of learned grasps by visualizing 2D histograms of successful grasps, mapped directly on the hammer object in Fig. \ref{fig:grasp-supervision}. To create this visualization from simulation, we record each grasping position by saving the 3D location (with respect to the hammer) of the middle point between gripper fingertips after each successful grasp. We then project the grasping positions recorded over 15,000 training steps onto a 2D histogram, where darker regions indicate more grasps. The silhouette of the hammer is outlined in black, with a green dot indicating its CoM. We illustrate the grasp histograms of three policies: Residual-physics with grasping supervised by heuristic that checks grasp width after grasping primitive (left), Residual-physics with grasping supervised by accuracy of throws (middle), and Physics-only with grasping supervised by accuracy of throws (right).

The differences between left and middle histograms indicate that leveraging accurate throws as a supervisory signal encourages the grasping policy to learn a more restricted but stable and homogeneous set of grasps: slightly further from the CoM to avoid unintentional collisions between the fingers and rest of the object at the moment of release, but also further from the ends of the handle to avoid less predictable throws. The differences between middle and right histograms show that when using only ballistics for the throwing module (\ie without learning throwing), the grasping policy seems to further optimize for grasps that are closer to the CoM. This leads to a more restricted set of grasps in contrast to Residual-physics, where the throwing can learn to compensate respectively. 

\begin{figure*}[t]
\centering
  \includegraphics[width=\linewidth]{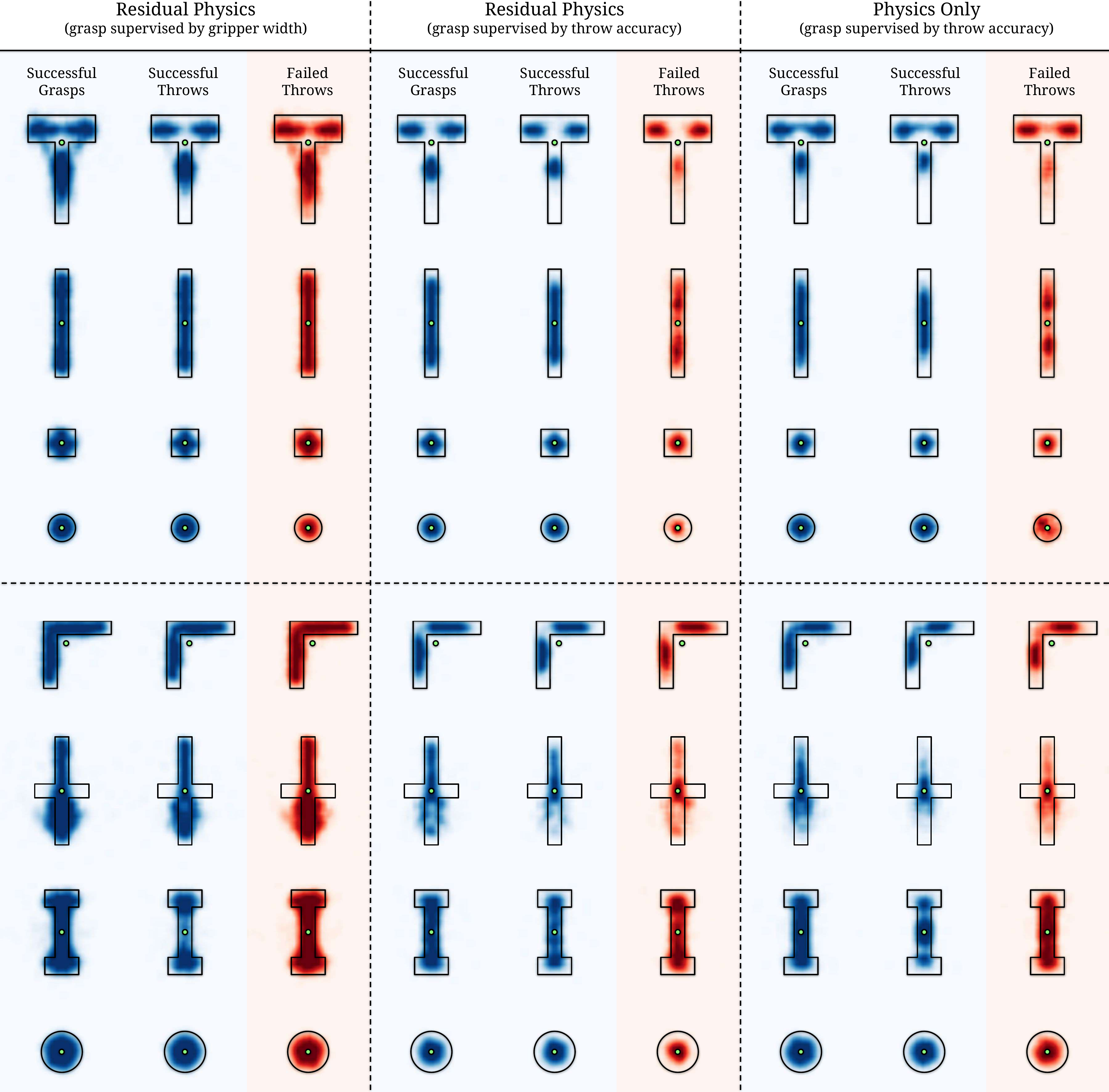}
  \caption{\textbf{Additional grasping histograms} of all simulation objects. Histograms are generated for successful grasps, grasps that lead to successful throws, and grasps that lead to failed throws -- recorded over 15,000 training steps. Darker regions indicate more grasps. The silhouette of each object is outlined in black, with a green dot indicating its CoM.}
  \label{fig:additional-grasp-histograms}
\end{figure*}

We also provide similar 2D grasp histogram visualizations in Fig. \ref{fig:additional-grasp-histograms} for all simulation objects. Across all policies, the histograms visualizing grasps which lead to successful throws (columns 2, 5, 8) share large overlaps with the grasps that lead to failed throws (red columns 3, 6, 9). This suggests grasping and throwing might have been learned simultaneously, rather than one after the other -- likely because the way the robot throws is conditioned on how it grasps in a non-trivial manner.



\subsection{Generalizing to New Target Locations} 

One of the key benefits from any Residual Physics approach is that the physics-based part of the controller naturally generalizes to conditions outside the collected data, for example to new target locations.
To explore how well our trained TossingBot policies generalize to new target locations, we displace the locations of the boxes in both the horizontal plane from where they were during training, such that there is no overlap between training and testing locations. For this experiment, we set in simulation 12 training boxes and 12 testing boxes; while in real settings, we set 4 training and 4 testing boxes (limited by physical setup). We record each model's throwing performance on seen objects over these new box locations across 1,000 testing steps in Table \ref{table:experiments-new-locations}.

\begin{table}[h]
  \centering
  \caption{Throwing to Unseen Locations (Mean \%)}
  \vspace{-1em}
  \begin{tabular}{l c c}
  \toprule
  Method & Simulation & Real \\  
  \midrule
  Regression-PoP    & 26.5 & 32.7 \\   
  Physics-only      & 79.6 & 62.2 \\    
  Residual-physics  & \bf{87.2} & \bf{83.9} \\  
  \bottomrule
  \end{tabular}
  \label{table:experiments-new-locations}
  \vspace{-1em}
\end{table} 

We observe that both in simulated and in real experiments,  Residual-physics significantly outperforms the regression baseline. The performance margin in this scenario illustrates how Residual-physics leverages the generalization of the ballistic equations to adapt to new target locations.

\subsection{Deep Object Semantics Emerging from Task Training}
\label{sec:emerging-semantics}

In this section, we explore the deep features being learned by the neural network $f$ and answer the questions: ``What does TossingBot learn from grasping and throwing arbitrary objects?''  and ``Do they convey any meaningful structure or representation?'' We do this by analyzing how similarly or differently the learned network handles different objects.

\begin{figure*}[t]
  \centering
  \includegraphics[width=\linewidth]{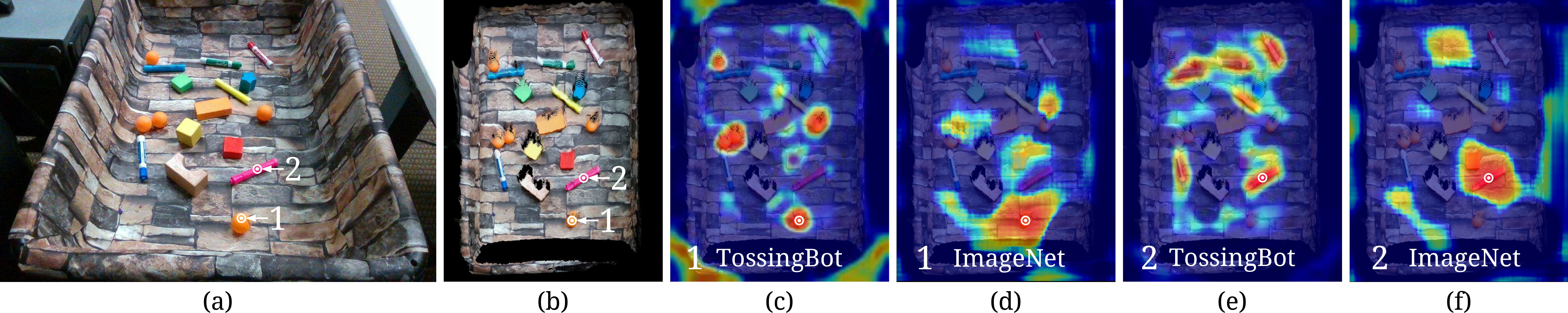}
  \caption{\textbf{Emerging semantics from interaction.} Visualizing pixel-wise deep features $\mu$ learned by TossingBot (c,e) overlaid on the input heightmap image (b) generated from an RGB-D side-view (a) of a bin of objects. (c) shows a heatmap of pixel-wise feature distances (hotter = smaller distance) from the feature vector of a query pixel on a ping pong ball (labeled 1). Likewise, (e) shows a heatmap of pixel-wise feature distances from the feature vector of a query pixel on a pink marker pen (labeled 2). These visualizations show that TossingBot learns features that distinguish object categories from each other without explicit supervision (\ie only task-level grasping and throwing). For reference, the same visualization technique is used on deep features generated by a ResNet-18 pre-trained on ImageNet (d,f).}
  \label{fig:emerging-semantics}
  \vspace{-1em}
\end{figure*}

To this end, we place several training objects in the bin (well-isolated from each other for visualization purposes), capture RGB-D images to construct a heightmap $I$ of the scene, and feed it through the network $f$ (already trained for 15,000 steps from real experiments). The training objects include marker pens, ping pong balls, and wooden toy blocks of different shapes (see Fig. \ref{fig:emerging-semantics}). We then extract the intermediate spatial feature representation of the network $\mu$ (described in Sec. \ref{sec:visual-representation}), which contains a 512-dimensional feature vector for each pixel of the heightmap $I$ (after $4\times$ upsampling to the same resolution). We then extract the feature vector from a query pixel belonging to one of the objects in the bin (ping pong ball in this case), and visualize its distance to all other pixel-wise features as a heatmap in Fig. \ref{fig:emerging-semantics}a (where hotter regions indicate smaller distances), overlaid on the original input heightmap. More specifically, we rank each pixel based on its $\ell_2$ feature distance to the query pixel feature, then colorize it based on its rank (\ie higher rank = closer feature distance = hotter color).

The procedure creates a form of similarity map between pixels. Interestingly, when choosing the pixel from a ping pong ball, the visualization immediately localizes all other ping pong balls in the scene -- presumably because they share similar deep features. It is also interesting to note that the orange wooden block, despite sharing a similar color, does not get picked up by the query. Similarly, Fig. \ref{fig:emerging-semantics}b illustrates the feature distances between a query pixel on a pink marker pen to all other pixels of the scene. The visualization immediately localizes all other marker pens, which share similar shape and mass, but do not necessarily share color textures.

These interesting results suggest that the deep network is learning to bias the features (\ie learning a prior) based on the objects' shapes more so than their visual textures or color. The network likely learns that geometric cues are more useful for learning grasping and throwing policies -- \ie provides more information related to grasping interactions and projectile behaviors. In addition to shape, one could also argue that the learned deep features reflect the second-order (beyond visual or geometric) physical attributes of objects which influence their aerial behaviors when thrown. This perspective is also plausible, since the throwing policies are effectively learning to compensate for these physical attributes respectively. For comparison, these visualizations generated by features from TossingBot are more informative in this setting than those generated using deep features from a 18-layer ResNet pre-trained on ImageNet (also shown in Fig. \ref{fig:emerging-semantics}).

These emerging features were learned implicitly from scratch without any explicit supervision beyond task-level grasping and throwing. Yet, they seem to be sufficient for enabling the system to distinguish between ping pong balls and markers. As such, this experiment speaks out to a broader concept related to machine vision: how should robots learn the semantics of the visual world? From the perspective of classic computer vision, semantics are often pre-defined using human-fabricated image datasets and manually constructed class categories (\ie this is a ``hammer'', and this is a ``pen''). However, our experiment suggests that it is possible to implicitly learn such object-level semantics from physical interactions alone (as long as they matter for the task at hand). The more complex these interactions, the higher the resolution of the semantics. Towards more generally intelligent robots -- perhaps it is sufficient for them to develop their own notion of semantics through interaction \cite{xu2019densephysnet}, without human guidance.

\section{Discussion and Future Work}

This paper presents a framework for jointly learning grasping and throwing policies that enable TossingBot to pick-and-throw arbitrary objects from an unstructured bin into boxes located outside its maximum reach range at 500+ MPPH. We show that a key is the use of {\em Residual Physics}, a hybrid controller that leverages deep learning to predict residuals on top of control parameters estimated with physics. The combination enables the data-driven predictions to focus on learning the aspects of dynamics that are difficult to model analytically. Our experiments in both simulation and real settings show that the system: 1) learns to improve grasps for throwing through joint training from trial and error, and 2) performs significantly better with Residual Physics than comparable alternatives. 

The proposed system is a prototype with several limitations that suggest directions for future work. First, it assumes that objects are rigid and robust enough to withstand forces encountered when thrown -- further work is required to train networks to predict motions that account for fragile, articulated, or deformable objects. Second, it infers object-centric properties and dynamics only from visual data (an RGB-D image of the bin) -- exploring additional sensing modalities such as force-torque or tactile may enable the system to better react to new objects and better adapt its throwing velocities. Third, it is only able to infer the parameters needed to get an object to land in a target location -- it would be interesting to explore how to achieve more fine-grained control of the pose (including orientation) of an object in flight, potentially to reach a target landing pose while avoiding or leveraging external obstacles. Finally, we have so far demonstrated the benefits of Residual Physics only in the context of throwing -- investigating how the idea generalizes to other tasks is a promising direction for future research.




\section*{Acknowledgments}

We thank Ryan Hickman for managerial support, Ivan Krasin and Stefan Welker for technical discussions, Brandon Hurd, Julian Salazar and Sean Snyder for hardware support, Chad Richards and Jason Freidenfelds for feedback on writing, Erwin Coumans for advice on PyBullet, Laura Graesser for video narration, and Regina Hickman for photography. We are also grateful for hardware and financial support from Google, Amazon, Intel, NVIDIA, ABB Robotics, and Mathworks.



\footnotesize
\bibliographystyle{plainnat} 
\bibliography{main.bib}

\clearpage
\normalsize


%
\begin{IEEEbiography}[{\includegraphics[width=1ine,height=1.25in,clip,keepaspectratio]{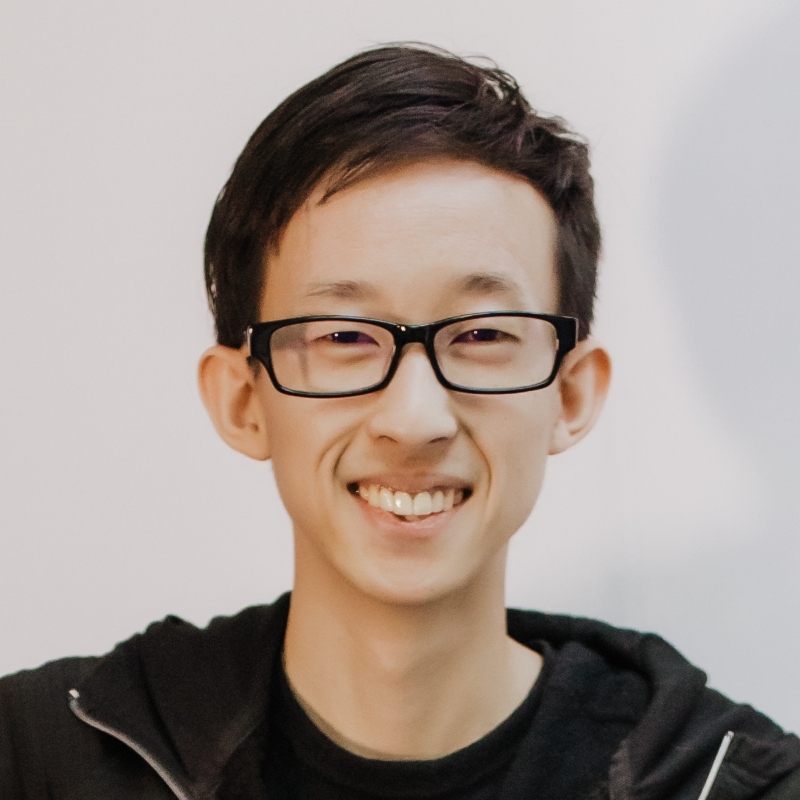}}]{Andy Zeng} is a research scientist in Robotics at Google AI. His research interests lie in vision for robotics -- developing learning algorithms that enable machines to intelligently interact with the physical world and improve themselves over time. Andy received his Bachelors in Computer Science ('15) and Mathematics ('15) at UC Berkeley, and earned his PhD in Computer Science ('19) at Princeton University. Andy received the Best System Paper Award at RSS ('19), Best Manipulation System Paper Award from Amazon ('18), and finalist for Best Cognitive Robotics Paper Award at IROS ('18). His research has been recognized through the Gordon Y.S. Wu Fellowship in Engineering and Wu Prize ('16), NVIDIA Fellowship ('18), and Princeton SEAS Award for Excellence ('18).
\end{IEEEbiography}

\begin{IEEEbiography}[{\includegraphics[width=1ine,height=1.25in,clip,keepaspectratio]{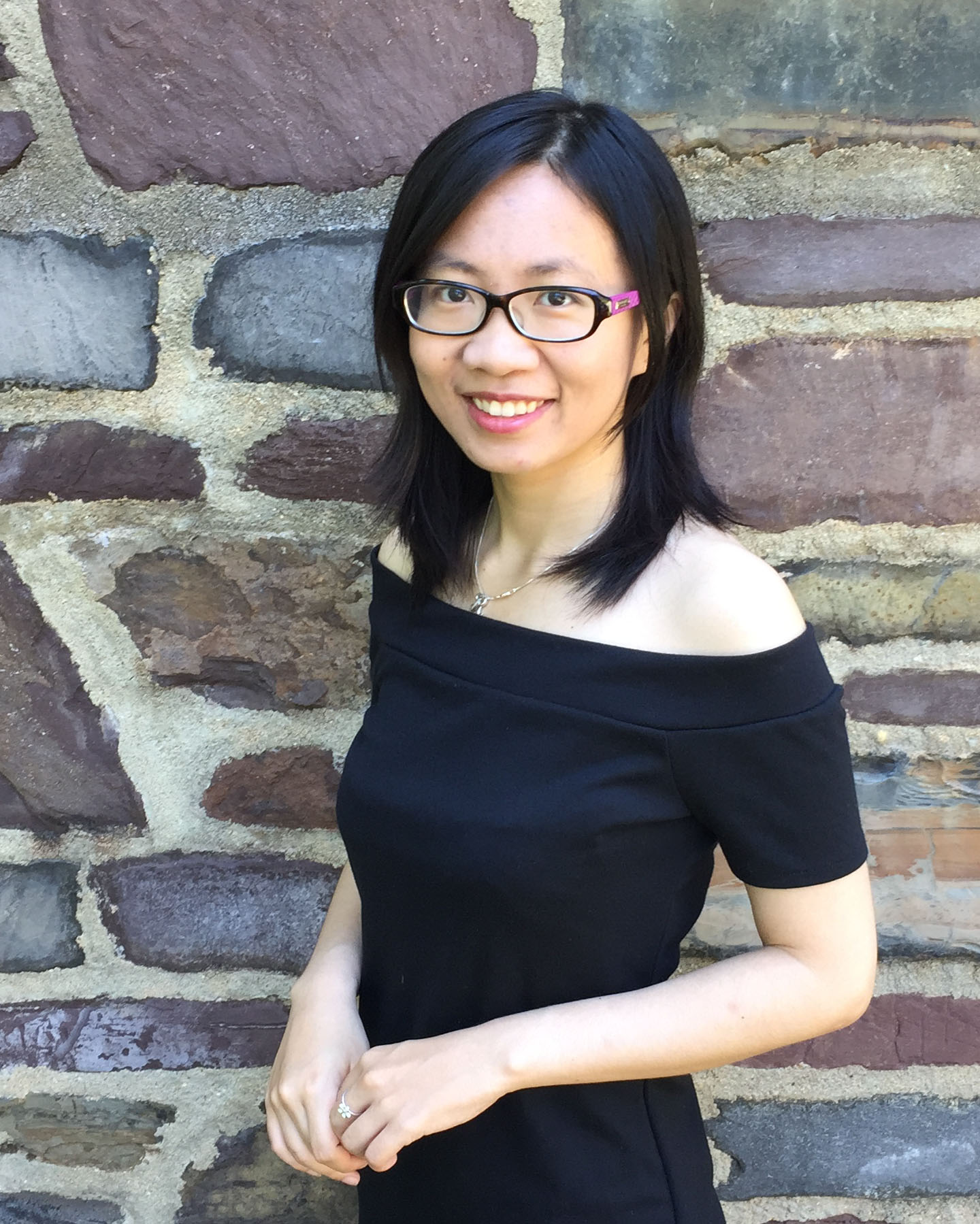}}]{Shuran Song} is an assistant professor in the Department of Computer Science at Columbia University. Before that, she received her Ph.D. in Computer Science at Princeton University, BEng. at HKUST in 2013. During her Ph.D., she spent time working at Microsoft Research and Google Daydream. Her research interests lie at the intersection of computer vision, robotics, and computer graphics.  She received the RSS Best System Paper Award in 2019, Facebook Fellowship in 2014, Siebel Scholar in 2016, Wallace Fellowship in 2017, and Princeton SEAS Award for Excellence in 2017.
\end{IEEEbiography}

\begin{IEEEbiography}[{\includegraphics[width=1ine,height=1.25in,clip,keepaspectratio]{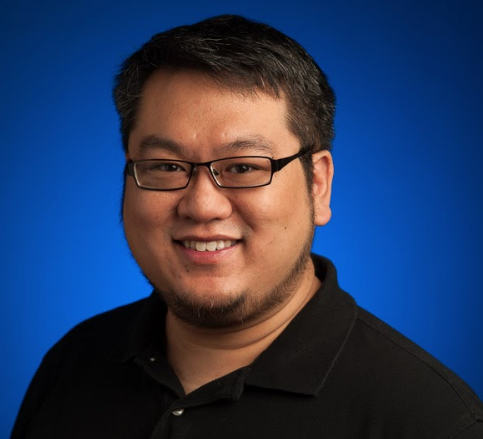}}]{Johnny Lee} is an Engineering Director in Robotics at Google, which is a division focused on applying machine learning to the field of robotics. His areas of interest include machine learning, computer vision, sensors, robotics, and human-computer interaction. Previously, he led the development of hardware and software technologies for AR/VR such as real-time motion tracking and environment mapping, and early stage R\&D efforts in Tango/Daydream. Previously, he helped Google X explore new projects as a Rapid Evaluator, and was a core algorithms contributor to the original Xbox Kinect. His YouTube videos demonstrating Wii remote hacks have surpassed over 15 million views and became one of the most popular TED talk videos. In 2008, he received his PhD in Human-Computer Interaction from Carnegie Mellon University and has been recognized in MIT Technology Review's TR35.
\end{IEEEbiography}

\begin{IEEEbiography}[{\includegraphics[width=1ine,height=1.25in,clip,keepaspectratio]{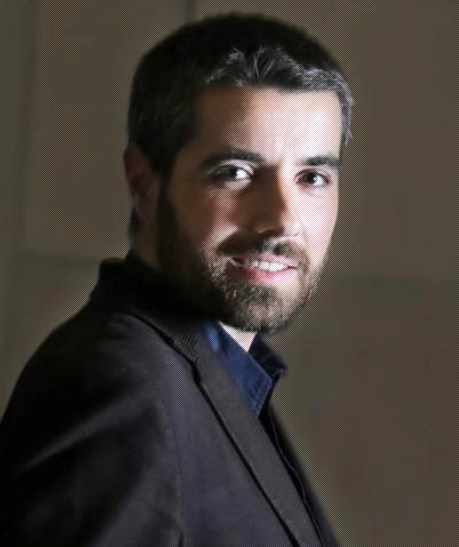}}]{Alberto Rodriguez} is an Associate Professor at the Mechanical Engineering Department at MIT. Alberto graduated in Mathematics ('05) and Telecommunication Engineering ('06) from the Universitat Politecnica de Catalunya, and earned his PhD (’13) from the Robotics Institute at Carnegie Mellon University. He leads the Manipulation and Mechanisms Lab at MIT (MCube) researching autonomous dexterous manipulation, robot automation, and end-effector design. Alberto has received Best Paper Awards at conferences RSS’11, ICRA’13, RSS’18, IROS'18, and RSS'19, the 2018 Best Manipulation System Paper Award from Amazon, and has been finalist for best paper awards at conferences IROS’16 and IROS'18. He has lead Team MIT-Princeton in the Amazon Robotics Challenge between 2015 and 2017, and has received Faculty Research Awards from Amazon in 2018, 2019 and 2020, and from Google in 2020. He is also the recipient of the 2020 IEEE Early Academic Career Award in Robotics and Automation. 
\end{IEEEbiography}

\begin{IEEEbiography}[{\includegraphics[width=1ine,height=1.25in,clip,keepaspectratio]{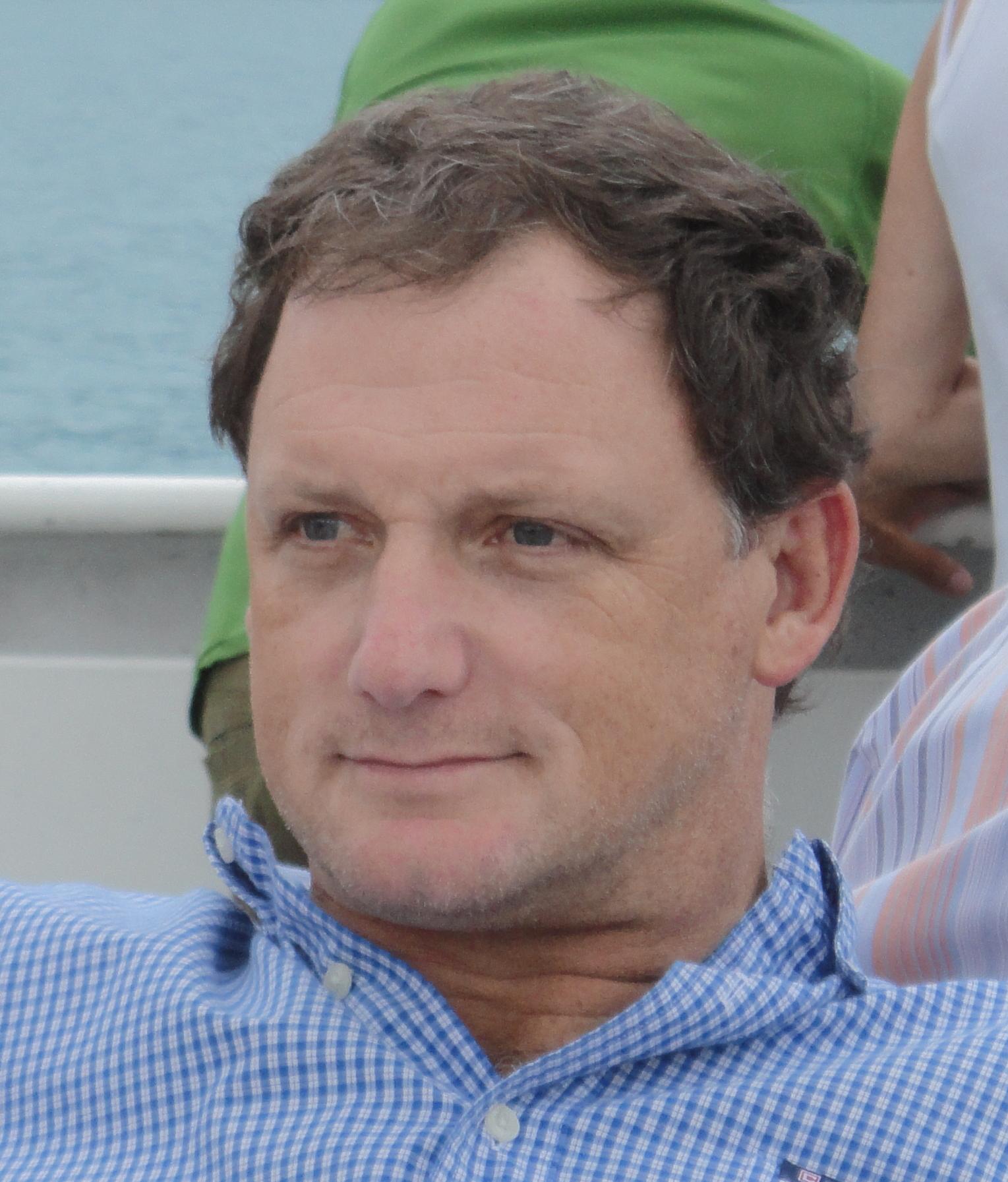}}]{Thomas Funkhouser} is a senior research scientist at Google and the David M. Siegel Professor of Computer Science, Emeritus, at Princeton University.  He received a PhD in computer science from UC Berkeley in 1993, worked as a member of the technical staff at Bell Labs for four years, and then taught at Princeton for twenty years.  His research interests include computer graphics, computer vision, and robotics.
\end{IEEEbiography}
\end{document}